\documentclass[10pt,twocolumn,letterpaper]{article}

\usepackage{iccv}
\usepackage{times}
\usepackage{epsfig}
\usepackage{graphicx}
\usepackage{amsmath}
\usepackage{amssymb}
\usepackage{bm} 

\usepackage{booktabs}
\usepackage{multirow}
\usepackage{algorithm}
\usepackage{algorithmic}
\renewcommand{\algorithmiccomment}[1]{\bgroup\hfill$\triangleright$~#1\egroup}

\usepackage[pagebackref=true,breaklinks=true,letterpaper=true,colorlinks,bookmarks=false]{hyperref}

\usepackage[title]{appendix}

\iccvfinalcopy 

\def\iccvPaperID{7438} 

\ificcvfinal\fi

\begin{document}

\title{Hypernetwork-Based Augmentation}

\author{
  Chih-Yang Chen and Che-Han Chang \\
  HTC Research \& Healthcare \\
}

\maketitle
\ificcvfinal\thispagestyle{empty}\fi

\newcommand{\ignore}[1]{}   
\newcommand{\cmt}[1]{\begin{sloppypar}\large\textcolor{red}{#1}\end{sloppypar}}
\newcommand{\note}[1]{\cmt{Note: #1}}

\newcommand{\todo}[1]{ \textcolor{red}{[{\bf TODO}: #1]}}
\newcommand{\torevise}[1]{\textcolor{blue}{#1}}
\newcommand{\copied}[1]{ \textcolor{red}{[COPIED: #1]}}
\newcommand{\frank}[1]{\textcolor{blue}{[Frank: #1]}}
\newcommand{\ed}[1]{\textcolor{blue}{[Ed: #1]}}
\newcommand{\harry}[1]{\textcolor{blue}{[harry: #1]}}

\newcommand{\comment}[1]{\textcolor{blue}{#1}}
\newcommand{\revised}[1]{\textcolor{green}{#1}}

\newcommand{\blue}[1]{\textcolor{blue}{#1}}
\newcommand{\ascalar}[1]{#1}
\newcommand{\avector}[1]{{\mathbf{#1}}}
\newcommand{\amatrix}[1]{\mathbf{#1}}
\newcommand{\aset}[1]{\mathbf{#1}}

\newcommand{\V}{\surd}

\newcommand{\stnPhi}{\bm{\phi}}
\newcommand{\diag}[1]{\text{diag}(#1)}

\newcommand{\stnW}{\mathbf{W}}
\newcommand{\Whyper}{\stnW_{\text{hyper}}}
\newcommand{\Welem}{\stnW_{\text{elem}}}
\newcommand{\stnV}{\mathbf{V}}

\newcommand{\stnb}{\mathbf{b}}
\newcommand{\bhyper}{\stnb_{\text{hyper}}}
\newcommand{\belem}{\stnb_{\text{elem}}}
\newcommand{\stnC}{\mathbf{C}}

\newcommand{\cin}{c_{1}}
\newcommand{\cout}{c_{2}}
\newcommand{\hpN}{n}

\newcommand{\Real}{\mathbb{R}}

\newcommand{\hp}{\lambda}
\newcommand{\w}{\theta} 
\newcommand{\LV}{L_V}
\newcommand{\LT}{L_T}
\newcommand{\bestw}{\w^{*}}
\newcommand{\hnetw}{\phi}
\newcommand{\hnet}{\hat{\w}_{\hnetw}}
\newcommand{\mi}{i}
\newcommand{\mN}{n}
\newcommand{\mk}{k}
\newcommand{\trainstepN}{T_{train}}
\newcommand{\valstepN}{T_{val}}
\newcommand{\trainLR}{\alpha}
\newcommand{\valLR}{\alpha'}
\newcommand{\loss}{\mathcal{L}}
\newcommand{\model}{\mathcal{F}}
\newcommand{\aug}{\mathcal{A}}
\newcommand{\inputimg}{x}
\newcommand{\inputimgval}{\inputimg^{v}}
\newcommand{\gtlabel}{y}
\newcommand{\gtlabelval}{\gtlabel^{v}}
\newcommand{\valset}{D_V}
\newcommand{\trainset}{D_T}
\newcommand{\noisehp}{\epsilon}
\newcommand{\noisew}{\epsilon'}
\newcommand{\gaussian}{N}
\newcommand{\sigmahp}{\sigma}
\newcommand{\sigmaw}{\sigma'}
\newcommand{\outerN}{T}

\newcommand{\linear}{f}
\newcommand{\inputvec}{x}
\newcommand{\outputvec}{y}
\newcommand{\indim}{c_{1}}
\newcommand{\outdim}{c_{2}}

\newcommand{\weightmatrix}{W}
\newcommand{\hyperlinear}{\hat{\weightmatrix}}
\newcommand{\hyperlinearbias}{W_0}
\newcommand{\hyperlinearU}{U}
\newcommand{\hyperlinearV}{V}

\newcommand{\subhyperlinear}{\hat{w}}
\newcommand{\subhyperlinearbias}{w}
\newcommand{\subhyperlinearweight}{A}
\newcommand{\uvec}{u}
\newcommand{\vvec}{v}
\newcommand{\Umat}{U}
\newcommand{\Vmat}{V}

\newcommand{\bias}{b}
\newcommand{\hyperbias}{\hat{\bias}}
\newcommand{\hyperbiasbias}{\hnetw_0^{\text{b}}}
\newcommand{\hyperbiasU}{\hnetw_U^{\text{b}}}
\newcommand{\hyperbiasV}{\hnetw_V^{\text{b}}}

\newcommand{\convONEweight}{\w_\text{conv1x1}}
\newcommand{\hyperconvONE}{\hat{\w}_\text{conv1x1}}
\newcommand{\hyperconvONEbias}{\hnetw_0^\text{c}}
\newcommand{\hyperconvONEU}{\hnetw_U^\text{c}}
\newcommand{\hyperconvONEV}{\hnetw_V^\text{c}}

\newcommand{\convweight}{\w_\text{conv}}
\newcommand{\hyperconv}{\hat{\w}_\text{conv}}
\newcommand{\hyperconvbias}{\hnetw_0^\text{c}}
\newcommand{\hyperconvU}{\hnetw_U^\text{c}}
\newcommand{\hyperconvV}{\hnetw_V^\text{c}}

\newcommand{\BNaffine}{\w_\text{BN}}
\newcommand{\hyperBNbias}{\hnetw_0^\text{BN}}
\newcommand{\hyperBNU}{\hnetw_U^\text{BN}}
\newcommand{\hyperBNV}{\hnetw_V^\text{BN}}

\begin{abstract}
Data augmentation is an effective technique to improve the generalization of deep neural networks.
Recently, AutoAugment~\cite{cubuk2019autoaugment} proposed a well-designed search space and a search algorithm that automatically finds augmentation policies in a data-driven manner.
However, AutoAugment is computationally intensive.
In this paper, we propose an efficient gradient-based search algorithm, called Hypernetwork-Based Augmentation (HBA), which simultaneously learns model parameters and augmentation hyperparameters in a single training.
Our HBA uses a hypernetwork to approximate a population-based training algorithm, which enables us to tune augmentation hyperparameters by gradient descent.
Besides, we introduce a weight sharing strategy that simplifies our hypernetwork architecture and speeds up our search algorithm.
We conduct experiments on CIFAR-10, CIFAR-100, SVHN, and ImageNet.
Our results show that HBA is competitive to the state-of-the-art methods in terms of both search speed and accuracy.
\end{abstract}

\section{Introduction}

Data augmentation techniques, such as cropping, horizontal flipping, and color jittering, are widely used in training deep neural networks for image classification.
Data augmentation acts as a regularizer that reduces overfitting by transforming images to increase the quantity and diversity of training data.
Recently, data augmentation has shown to be an effective technique not only for supervised learning~\cite{cubuk2019autoaugment, ho2019population, lim2019fast}, but also for semi-supervised learning~\cite{xie2019unsupervised,berthelot2019mixmatch}, self-supervised learning~\cite{chen2020simple}, and reinforcement learning (RL)~\cite{kostrikov2020image}.
However, given a new task or dataset, it is non-trivial to either manually or automatically design its data augmentation policy: determine a set of augmentation functions and adequately specify their configurations, such as the range of rotation, the size of cropping, the degree of color jittering. A weak augmentation policy may not help much, while a strong one could hinder performance by making an augmented image looks inconsistent to its label.

AutoAugment~\cite{cubuk2019autoaugment}, a representative pioneering work proposed by Cubuk et al., proposed a search space (consisting of 16 image operations) and an RL-based search algorithm to automate the process of finding an effective augmentation policy over the search space.
AutoAugment made remarkable improvements in image classification. However, its search algorithm requires $5000$ GPU hours for one augmentation policy learning, which is extremely computationally intensive. 

\begin{table}
\small
\centering
\label{table:teaser}
\begin{tabular}{cccccc}
\toprule
Dataset &            & AA    & PBA & FAA  & HBA  \\
\midrule
\multirow{2}{*}{CIFAR-10} & GPU Hours  & $5000$  & $5$   & $3.5$  & $0.1$  \\
                          & Test Error {\tiny $(\%)$} & $2.6$   & $2.6$ & $2.7$  & $2.6$  \\
\midrule
\multirow{2}{*}{SVHN} & GPU Hours  & $1000$  & $1$   & $1.5$  & $0.1$  \\
                          & Test Error {\tiny $(\%)$} & $1.1$   & $1.2$ & $1.1$  & $1.1$  \\
\midrule
\multirow{2}{*}{ImageNet} & GPU Hours  & $15000$ & -     & $450$  & $0.9$  \\
                          & Test Error {\tiny $(\%)$} & $22.4$  & -     & $22.4$ & $22.1$ \\
\bottomrule
\end{tabular}
\vspace{2pt}
\caption{
Hypernetwork-Based Augmentation (HBA) is at least an order of magnitude faster than AutoAugment (AA)~\cite{cubuk2019autoaugment}, Population Based Augmentation (PBA)~\cite{ho2019population}, and Fast AutoAugment (FAA)~\cite{lim2019fast}, while achieving similar accuracy.
}
\vspace{-12pt}
\end{table}

To address the computational issue, we formulate the augmentation policy search problem as a hyperparameter optimization problem and propose an efficient algorithm for tuning data augmentation hyperparameters.
The central idea of our method is a novel combination of a population-based training algorithm and a \textit{hypernetwork}, which is a function that outputs the weights of a neural network.
In particular, we introduce a population-based training procedure and four modifications to derive our algorithm.
First, we employ a hypernetwork to represent the set of models.
By doing so, training a hypernetwork can be treated as training a continuous population of models.
Second, we propose a new hyper-layer for batch normalization~\cite{ioffe2015batch} to facilitate the construction of hypernetworks.
Third, inspired by one-shot neural architecture search~\cite{pham2018efficient, brock2017smash, bender2018understanding}, we adopt a weight sharing strategy to the models.
We show that such a strategy corresponds to a simplified hypernetwork architecture that effectively reduces the search time.
Fourth, instead of evaluating the performance on the discrete set of models, thanks to the use of a hypernetwork, we perform gradient descent to efficiently find the approximate best model.
Our method, dubbed as Hypernetwork-Based Augmentation (HBA), is a \textit{gradient-based} method that jointly trains the network model and tunes the augmentation hyperparameters.
HBA yields hyperparameter schedules that can be used to train on different datasets or to train different network architectures.
Experimental results show that HBA achieves significantly faster search speed while maintaining competitive accuracy.
Table~\ref{table:teaser} summarizes our main results.





Our contributions can be summarized as follows.
(1) We propose Hypernetwork-Based Augmentation (HBA), an efficient gradient-based method for automated data augmentation.
(2) The derivation of HBA reveals the underlying relationship between population-based and gradient-based methods.
(3) We propose a weight-sharing strategy that simplifies our hypernetwork architecture and reduces the search time considerably without performance drop.







\section{Related work}

\begin{figure*}
    \centering
    \begin{tabular}{c}
    \includegraphics[width=0.98\linewidth]{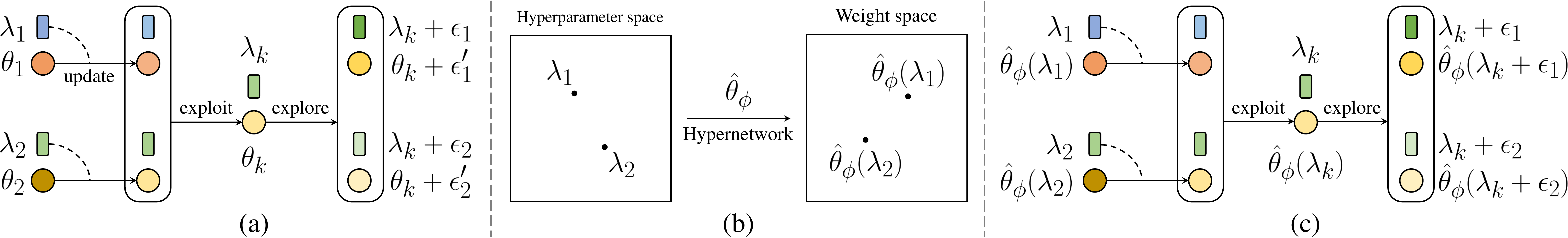}
    \end{tabular}
    \caption{(a) A population-based training (PBT) trains a discrete set of models and performs exploitation followed by exploration to tune data augmentation hyperparameters. In this example, two models are updated, and the second one is exploited and explored. Section~\ref{3.1} explains the details.
    (b) We employ a hypernetwork $\hnet$ to represent the set of models $\{\w_1, \w_2\}$ by $\{\hnet(\hp_1), \hnet(\hp_2)\}$, where $\hnet$ is a continuous function that aims to map from each model $\mi$'s hyperparameters $\hp_\mi$ to its weights $\w_\mi$.
    (c) We show the resulting schematic diagram of training a single hypernetwork that approximates a population-based training. Section~\ref{3.2} details the modified training algorithm, and Section~\ref{3.3} describes the hypernetwork architecture.
    }
    \label{fig:pbt_by_hypernet}
\end{figure*}

\noindent\textbf{Data Augmentation.} We review previous work relevant to the development of our method.
We refer readers to a comprehensive survey paper~\cite{shorten2019survey} on data augmentation.
Hand-designed data augmentation techniques, such as horizontal flipping, random cropping, and color transformations, are commonly used in training deep neural networks for image classification~\cite{krizhevsky2012imagenet,he2016deep}.
In recent years, several effective data augmentation techniques are proposed.
Cutout~\cite{devries2017improved} randomly erases contents by sampling a patch from the input image and replace it with a constant value.
Mixup~\cite{zhang2017mixup} performs data interpolation that combines pairs of images and their labels in a convex manner to generate virtual training data.
CutMix~\cite{yun2019cutmix} generates new samples by cutting and pasting image patches within mini-batches.
These data augmentation techniques are hand-designed, and their hyperparameters are usually manually-tuned.

\noindent\textbf{Automated Data Augmentation.} Inspired by recent advances in neural architecture search~\cite{ zoph2017neural, pham2018efficient, liu2019darts}, one of the recent trends in data augmentation is automatically finding augmentation policies within a pre-defined search space in a data-driven manner.
AutoAugment~\cite{cubuk2019autoaugment} introduced a well-designed search space and proposed an RL-based search algorithm that trains a recurrent neural network (RNN) controller to search effective augmentation policies within the search space.
Although AutoAugment achieves promising results, its search process is computationally expensive (5000 GPU hours on CIFAR-10).
Our algorithm treats the data augmentation hyperparameters as continuous variables, and efficiently optimizes them by gradient descent in a single round of training.

Recently, several efficient search algorithms based on AutoAugment’s search space have been proposed.
Population Based Augmentation (PBA)~\cite{ho2019population} employed Population Based Training (PBT)~\cite{jaderberg2017population}, an evolution-based hyperparameter optimization algorithm, to search data augmentation schedules.
Fast AutoAugment~\cite{lim2019fast} treats the problem as a density matching problem and used Bayesian optimization to find augmentation policies.
OHL-Auto-Aug~\cite{ lin2019online} proposed an online hyperparameter learning algorithm that jointly learns network parameters and augmentation policy.
Our algorithm also tunes augmentation hyperparameters in an online manner while achieving better search efficiency.
RandAugment~\cite{cubuk2019randaugment} simplified the search space of AutoAugment and used the grid search method to find the optimal augmentation policy.
Differentiable Automatic Data Augmentation (DADA)~\cite{li2020dada}, inspired by Differentiable Architecture Search (DARTS)~\cite{liu2019darts}, proposed a gradient-based method that relaxes the discrete policy selection to be differentiable and optimize the augmentation policy by stochastic gradient descent.
Our algorithm is also gradient-based but is fundamentally different from DADA.
DADA realizes relaxation by the Gumbel-softmax trick, while ours is based on combining population-based training and hypernetworks.


\noindent\textbf{Hypernetworks.}
Ha et al.~\cite{ha2017hypernetworks} used hypernetworks to generate weights for recurrent networks.
SMASH~\cite{brock2017smash} presented a gradient-based neural architecture search method that employs a hypernetwork to learn a mapping from a binary-encoded architecture space to the weight space.
Self-Tuning Network (STN)~\cite{mackay2019self} used a hypernetwork as an approximation to the best response function in bilevel optimization.
Our method uses a hypernetwork to represent a set of models trained with different hyperparameters.


\noindent\textbf{Hyperparameter Optimization.} Searching augmentation policies can be formulated as a hyperparameter optimization problem.
We refer readers to a recent survey paper~\cite{yu2020hyper} on hyperparameter optimization.
Population Based Training (PBT)~\cite{ho2019population} presented a hyperparameter optimization algorithm that trains a population of models in parallel, and periodically evaluates their performance to perform exploitation and exploration.
MacKay et al. proposed the Self-Tuning Network (STN)~\cite{mackay2019self}, a gradient-based method for tuning regularization hyperparameters, including data augmentation and dropout~\cite{srivastava2014dropout}.

HBA is closely related to PBT~\cite{jaderberg2017population} and STN~\cite{mackay2019self}.
In our method, we start from a population-based training algorithm and apply a series of modifications to derive our algorithm, which can be viewed as an efficient approximation of the PBT algorithm.
Compared with STN, HBA distinguishes from STN in three aspects.
First, the underlying formulations are different. STN is based on the best response approximation to bilevel optimization, while HBA is derived entirely from the perspective of population-based training.
Second, HBA adopts a weight sharing strategy, which is intuitive yet novel since it cannot be analogously defined in STN.
Third, to facilitate the construction of hypernetworks, we propose a new hyper-layer for batch normalization. 





\section{Method}
\ignore{
Our proposed method HBA consists of two components: a search space and a search algorithm.
We describe our search space in Section~\ref{3.1} and our search algorithm in Section~\ref{3.2},~\ref{3.3},~\ref{3.4},~\ref{3.5} and~\ref{3.6}.
}
We start from a population-based training (Section~\ref{3.1}) and then present a series of modifications 
to derive HBA: hypernetworks (Section~\ref{3.2} and~\ref{3.3}), weight sharing strategy (Section~\ref{3.4}), and gradient-based exploitation (Section~\ref{3.5}). 


\subsection{Training a Population of Models by Stochastic Gradient Descent}  \label{3.1}

We formulate tuning data augmentation hyperparameters as training a population of $\mN$ models.
Each model $\mi$ has the same network architecture but a different initialization of both model parameters $\w_\mi$ and augmentation hyperparameters $\hp_\mi$.
During training, we periodically {\em exploit} the best performing model and {\em explore} both its augmentation hyperparameters and model parameters.
Specifically, our population-based training procedure iterates among three steps:
\textbf{update}, \textbf{exploit}, and \textbf{explore}. In the following, we detail each of them.

\noindent\textbf{Update.}
For each model $\mi$, we update its parameters $\w_\mi$ by $\trainstepN$ stochastic gradient descent (SGD) steps.
The formula of an SGD step can be expressed as
\begin{align}
    \w_\mi = \w_\mi - \trainLR \nabla_{\w} \loss(\model(\aug(\inputimg;\hp_\mi);\w_\mi), \gtlabel) \label{eq:sgd_bs1},
\end{align}
where $(\inputimg, \gtlabel)$ is a training example, $\model(; \w_i)$ is the $\mi$-th model, $\aug(; \hp_\mi)$ is the augmentation policy used for model $\mi$, $\loss$ is the loss function, and $\trainLR$ is the learning rate.

\noindent\textbf{Exploit.}
Our exploitation strategy evaluates each model $\mi$ on the validation set $\valset$, identifies the best performing model $\mk$, and copies its parameters and hyperparameters to replace the ones of all the other models. Specifically,
\begin{align}
    \hp_\mi &= \hp_\mk \text{~and} \\
    \w_\mi &= \w_\mk,
\end{align}
where
\begin{align}
    \mk = \arg\min_{\mi\in\{1,...,\mN\}} \sum_{(\inputimgval,\gtlabelval)\in \valset}\loss(\model(\inputimgval;\w_\mi), \gtlabelval). \label{eq:exploit0}
\end{align}

\noindent\textbf{Explore.}
We perform exploration by perturbing the value of each model's parameters and hyperparameters as
\begin{align}
    \hp_\mi &= \hp_\mi + \noisehp_\mi \text{~and} \label{eq:explore_hp}\\
    \w_\mi &= \w_\mi + \noisew_\mi, \label{eq:explore_w}
\end{align}
where $\noisehp_\mi \sim \gaussian(0,\sigmahp)$ and $\noisew_\mi \sim \gaussian(0,\sigmaw)$ are Gaussian noise.

Figure~\ref{fig:pbt_by_hypernet} (a) shows the schematic diagram of our population-based training.
Until now, we have presented a population-based training procedure, which serves as the foundation of HBA.
In the next two subsections, we introduce hypernetworks, which is the key component to convert the population-based training into a gradient-based training.

\subsection{Representing a Population of Models by a Hypernetwork}  \label{3.2}

We define a hypernetwork as a function that takes the data augmentation hyperparameters $\hp$ as inputs and outputs the parameters $\w$ of the neural network $\model$.
The hypernetwork itself is a neural network, and we denote it by $\hnet$ where $\hnetw$ is the hypernetwork parameters.
We use a hypernetwork to represent the members of the population by
\begin{align}
    \w_\mi = \hnet(\hp_\mi) \text{~~~~for~} \mi = 1,2,...,\mN. \label{eq:hnet}
\end{align}
Therefore, the set of model parameters $\{\w_\mi\}_{i=1}^{\mN}$ now are represented by a single hypernetwork $\hnet$.
Given model $\mi$'s hyperparameters $\hp_\mi$, its parameters $\w_\mi$ can be obtained through $\hnet$.
Figure~\ref{fig:pbt_by_hypernet} (b) illustrates the concept of adopting a hypernetwork to represent a set of models.


Based on Equation~\ref{eq:hnet}, training a population of models with parameters $\{\w_\mi\}_{\mi=1}^{\mN}$ is changed into training a single hypernetwork with parameters $\hnetw$.
The \textbf{update} step in Equation~\ref{eq:sgd_bs1} is accordingly changed as
\begin{align}
    \hnetw = \hnetw - \trainLR \frac{1}{\mN}\sum_{\mi=1}^{\mN} \nabla_{\hnetw}\loss(\model( \aug(\inputimg_\mi;\hp_\mi) , \hnet(\hp_\mi)), \gtlabel_\mi). \label{eq:sgd_bsn}
\end{align}
Comparing with Equation~\ref{eq:sgd_bs1} that independently updates each model parameters $\w_\mi$ by SGD with a batch size of $1$, Equation~\ref{eq:sgd_bsn} updates the hypernetwork parameters $\hnetw$ by SGD with a batch size of $\mN$.
Figure~\ref{fig:pbt_by_hypernet} (c) shows the schematic diagram of the modified population-based training that using a hypernetwork to represent the two models.


\begin{figure}
    \centering
    \begin{tabular}{c}
    \includegraphics[width=0.97\linewidth]{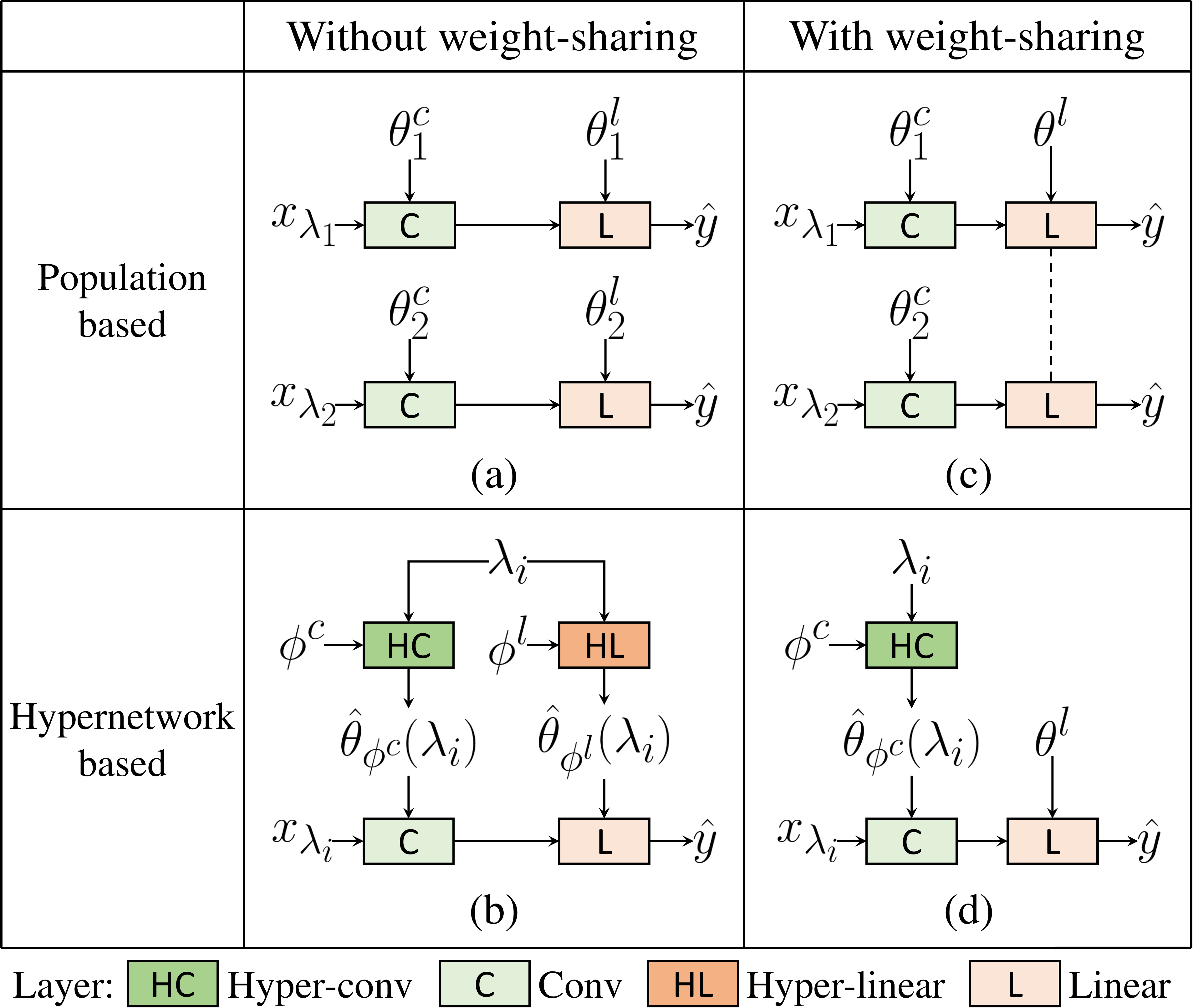}
    \end{tabular}
    \vspace{2pt}
    \caption{The model architecture of (a) population-based training, (b) hypernetwork-based training, (c) population-based training with weight sharing, and (d) hypernetwork-based training with weight sharing. We denote the augmented input images $\aug(\inputimg;\hp_1)$ and $\aug(\inputimg;\hp_2)$ by $\inputimg_{\hp_1}$ and $\inputimg_{\hp_2}$, respectively. Please see Section~\ref{3.3} and~\ref{3.4} for details.
    }
    \label{fig:hypernet_architecture}
\end{figure}

\subsection{Hypernetwork Architecture} \label{3.3}

Given the architecture of a neural network $\model$, we define its corresponding hypernetwork architecture as follows.
First, we construct a linear hypernetwork in a layer-wise manner: for each trainable layer in $\model$, we associate it with a hyper-layer accordingly, which is a linear layer.
Employing hypernetworks could be memory-intensive since even a linear hypernetwork requires a prohibitively large number of parameters, which would be $|\hp||\w|$.
Therefore, following STN~\cite{mackay2019self}, we assume that the weight matrix of each hyper-layer is low-rank to effectively reduce the parameter complexity from $O(|\hp||\w|)$ to $O(|\hp|+|\w|)$.

In our experiments, there are three types of trainable layers: convolutional (conv) layer, batch normalization (BN) layer, and fully-connected (linear) layer.
To associate each trainable layer with a hyper-layer,
we employ the hyper-conv and hyper-linear layer proposed by STN and follow its design spirit to propose a hyper-BN layer.

Our hyper-BN layer takes $\hp$ as input and outputs the affine parameters $\w_\text{BN} \in \Real^{c}$ of a BN layer by
$\w_\text{BN} = \text{HyperBN}(\hp;\hnetw_b,\hnetw_U,\hnetw_V) = \hnetw_b + \diag{\hnetw_V \hp}\hnetw_U$
, where $\hnetw_b,\hnetw_U \in \Real^{c}$, $\hnetw_V \in \Real^{c \times |\hp|}$, $c$ is the number of output channels, and $\diag{\cdot}$ turns a vector into a diagonal matrix.
\ignore{More details on the hyper-layers can be found in the supplementary material.}
\ignore{Due to the space limit, please refer to the supplementary material for the details of the hyper-layers.}

In Figure~\ref{fig:hypernet_architecture} (a), we show an example of a population-based training that trains two models, each of which consists of a conv layer and a linear layer.
We can convert the set of models into a single hypernetwork by attaching hyper-layers onto each trainable layer, as shown in Figure~\ref{fig:hypernet_architecture} (b).
Up to now, we have introduced hypernetworks and their architecture.
Next, we introduce weight sharing as a way to improve population-based training.

\begin{figure*}
    \centering
    \begin{tabular}{c}
    \includegraphics[width=0.98\linewidth]{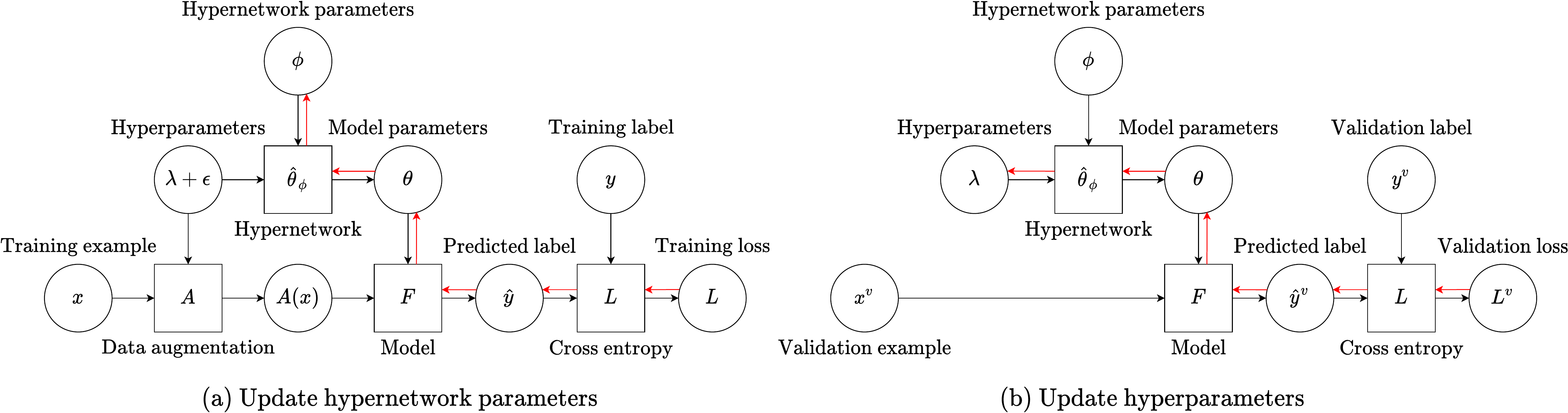}
    \end{tabular}
    \caption{The computation graph of the HBA algorithm. (a) We train a hypernetwork, which represents a population of models, to minimize the training loss (Equation~\ref{eq:sgd_w}). (b) We update the hyperparameters by minimizing the validation loss (Equation~\ref{eq:sgd_lambda}). The arrows in red represent the gradient flow in the backward propagation.
    }
    \label{fig:computation_graph}
\end{figure*}

\subsection{Weight Sharing across Models.} \label{3.4}

Weight sharing is a commonly-used strategy to promote cooperation in training multiple models.
Therefore, it looks natural to consider applying a weight sharing strategy in our population-based training.
Recall that the \textbf{update} step (Equation~\ref{eq:sgd_bs1}) independently updates each model in the population.
If some of the layers are shared among the population, the population members can be seen as performing a joint training in the \textbf{update} step.
As we can see from the example in Figure~\ref{fig:hypernet_architecture} (c),
if we share the weights of the linear layer,
these two models, trained with different hyperparameters, learn different convolutional features while sharing the same linear classifier.

Besides model cooperation, another reason that motivates us to apply weight sharing is based on the following observation:
applying a weight sharing strategy to the population is equivalent to adopting a \textit{simplified} hypernetwork architecture, which can effectively reduce both the search time and memory consumption.

For example, if all the population members share a particular layer's parameters, then it equivalently means that the corresponding hyper-layer degenerates into a \textit{constant function} independent of hyperparameters.
In other words, if we share a layer's parameters, there is no need to associate it with a hyper-layer.
By doing so, the number of hyper-layers becomes smaller, the memory consumption in the training process is reduced, and the search speed becomes faster.
Taking Figure~\ref{fig:hypernet_architecture} (d) as an example, if we simultaneously employ a hypernetwork and shares the linear layer, then only the conv layer is attached with a hyper-conv layer. The hyper-linear layer degenerates into a constant function and thus is removed, giving us a smaller hypernetwork than the one without weight sharing (Figure~\ref{fig:hypernet_architecture} (b)).
In summary, Figure~\ref{fig:hypernet_architecture} shows the corresponding model architectures of whether employing a hypernetwork and whether applying a weight sharing strategy.

\subsection{Gradient-based Exploitation} \label{3.5}

With hypernetwork $\hnet$, the \textbf{exploit} step (Equation~\ref{eq:exploit0}) can be modified accordingly as 
\begin{align}
\mk = \arg\min_{\mi\in\{1,...,\mN\}} \sum_{(\inputimgval,\gtlabelval)\in \valset}\loss(\model(\inputimgval;\hnet(\hp_\mi)), \gtlabelval). \label{eq:naive_hnet_exploit}
\end{align}
However, a hypernetwork, as a continuous function, naturally represents not only $\mN$ models but also a continuous population of models.
Therefore, we propose to find the best performing model $\hnet(\hp^*)$ in the continuous population by minimizing the validation loss over \textit{hyperparameters} as
\begin{align}
\hp^* = \arg\min_{\hp} \sum_{(\inputimgval,\gtlabelval)\in \valset}\loss(\model(\inputimgval;\hnet(\hp)), \gtlabelval). \label{eq:exploit2}
\end{align}

We use gradient descent to approximately yet efficiently find $\hp^*$.
In particular, our \textbf{exploit} step is changed as applying $\valstepN$ steps of SGD to solve Equation~\ref{eq:exploit2}, and each step can be written as
\begin{align}
    \hp = \hp - \valLR \frac{1}{\mN}\sum_{\mi=1}^\mN \nabla_{\hp}\loss(\model(\inputimgval_\mi, \hnet(\hp)), \gtlabelval_\mi). \label{eq:sgd_lambda}
\end{align}


\noindent\textbf{Explore and Update.}
Based on Equation~\ref{eq:sgd_lambda}, the \textbf{explore}  step of $\hp_\mi$ (Equation~\ref{eq:explore_hp}) can be modified accordingly as 
\begin{align}
    \hp_\mi = \hp + \noisehp_\mi. \label{eq:explore_hp2}
\end{align}
Besides, we let the model parameters be implicitly exploited and explored by $\w_\mi = \hnet(\hp_\mi)=\hnet(\hp + \noisehp_\mi)$, which can be seen as an approximation of $\hnet(\hp) + \noisew_\mi$.

Furthermore, by substituting Equation~\ref{eq:explore_hp2} into Equation~\ref{eq:sgd_bsn}, we merge the \textbf{explore} and \textbf{update} step together and obtain the formula of updating the hypernetwork as

\begin{align}
    \hnetw = \hnetw - \trainLR \frac{1}{\mN}\sum_{\mi=1}^{\mN} \nabla_{\hnetw}\loss(\model(\aug(\inputimg_\mi; \hp+\noisehp_\mi); \hnet(\hp+\noisehp_\mi)), \gtlabel_\mi). \label{eq:sgd_w}
\end{align}
where $\noisehp_\mi \sim \gaussian(0,\sigmahp)$.
Lastly, instead of using the same perturbation noise $\{\noisehp_\mi\}_{\mi=1}^{\mN}$ across the $\trainstepN$ SGD steps of Equation~\ref{eq:sgd_w}, we re-sample $\{\noisehp_\mi\}_{\mi=1}^{\mN}$ for each iteration to enhance the sample diversity.
In other words, at each SGD step, we sample a different discrete set of models from the continuous population to update the hypernetwork.

\noindent\textbf{Summary.} Algorithm~\ref{alg:hba} summarizes our HBA algorithm, which is a gradient-based training algorithm that alternates between updating the hypernetwork (Equation~\ref{eq:sgd_w}) and updating the hyperparameters (Equation~\ref{eq:sgd_lambda}).
HBA builds upon the concept of the population-based training (Equation~\ref{eq:sgd_bs1} to~\ref{eq:explore_w}) while performs a gradient-based training (Equation~\ref{eq:sgd_lambda} and~\ref{eq:sgd_w}).
In Figure~\ref{fig:computation_graph}, we show the computation graph of the HBA algorithm.

\noindent
\begin{minipage}{0.47\textwidth}
\vspace{-10pt}
\begin{algorithm}[H]
\algsetup{linenosize=\small}
\caption{
The algorithm of HBA.}
\label{alg:hba}
\begin{algorithmic}[1]
    \STATE {\bfseries Input:} training set $\trainset$, validation set $\valset$, model network $\model(\cdot;\theta)$, hypernetwork $\hnet$, augmentation policy $\aug(\cdot;\hp)$, batch size $\mN$, number of steps $\outerN$, $\trainstepN$ and $\valstepN$, learning rates $\trainLR$ and $\valLR$, standard deviation $\sigmahp$.
    \STATE Initialize $\hnetw$ and $\hp$
    \FOR{$j=1$ {\bfseries to} $\outerN$}
        \FOR{$t=1$ {\bfseries to} $\trainstepN$}
            \STATE Sample $\{(\inputimg_\mi,\gtlabel_\mi)\}_{\mi=1}^\mN$ from $\trainset$
            \STATE $\{ \noisehp_\mi \}_{\mi=1}^{\mN} \sim \gaussian(0, \sigmahp)$
            \STATE $\hnetw = \hnetw - \trainLR \frac{1}{\mN}\sum_{\mi=1}^{\mN} \nabla_{\hnetw}\loss(\model( \aug(\inputimg_\mi;\hp+\noisehp_\mi), \hnet(\hp+\noisehp_\mi)), \gtlabel_\mi)$ \COMMENT{Update $\hnetw$ (Equation~\ref{eq:sgd_w})}
        \ENDFOR
        \FOR{$t=1$ {\bfseries to} $\valstepN$}
            \STATE Sample $\{(\inputimgval_\mi,\gtlabelval_\mi)\}_{\mi=1}^\mN$ from $\valset$
            \STATE $\hp = \hp - \valLR \frac{1}{\mN}\sum_{\mi=1}^\mN \nabla_{\hp}\loss(\model(\inputimgval_\mi, \hnet(\hp)), \gtlabelval_\mi)$ \COMMENT{Update $\hp$ (Equation~\ref{eq:sgd_lambda})}
        \ENDFOR
    \ENDFOR
    \STATE {\bfseries Output:} $\hnet(\hp)$
\end{algorithmic}
\end{algorithm}
\end{minipage}

\section{Experiments}

\noindent\textbf{Implementation Details.}
\ignore{Following PBA~\cite{ho2019population}, our search space consists of $15$ operations, where each has two copies.
Please see the supplementary material for the list of operations and their hyperparameters.}
Following STN~\cite{mackay2019self}, HBA trains the hypernetwork parameters using SGD with $T_{train}=2$ and optimizes the hyperparameters using Adam~\cite{kingma2014adam} with $T_{val}=1$.
We implemented HBA using the Pytorch~\cite{paszke2019pytorch} framework.
We measured the search time of HBA on an NVIDIA Tesla V100 GPU.
We reported the last epoch's performance after training for all of our results using the mean and standard deviation over five runs with different random seeds.
For policy evaluation, we scaled the length of the discovered schedules linearly if necessary.

\noindent\textbf{Search Space.}
We define a data augmentation policy as a stochastic transformation function constructed by a set of elementary image operations. Each operation is associated with a probability and a magnitude, which play the role of data augmentation hyperparameters.
For a fair comparison in Section~\ref{4.1} and~\ref{4.2}, our transformation function follows PBA~\cite{ho2019population}, which consists of $15$ operations and performs three steps:
(1) sampling an integer $K$ from a categorical distribution as the number of operations to be applied, where $K\sim$ \{0: $p$ = 0.2, 1: $p$ = 0.3, 2: $p$ = 0.5\}.
(2) sampling $K$ operations based on the operation probabilities, 
and (3) sequentially applying these sampled operations to the input image.
\ignore{Please refer to the supplementary material or PBA~\cite{ho2019population} for the details of the PBA search space.}
Figure~\ref{fig:schedule} (b) lists the $15$ operations, whose detailed description can be found in the supplementary material.



\subsection{Ablation Study} \label{4.1}


\begin{table}
\centering
\begin{tabular}{c|ccc}
\toprule
          & CIFAR-10 & CIFAR-100 & GPU \\
Strategy  & Val. Error (\%) & Val. Error (\%) & Hours \\
\midrule
Conv + BN & $2.85 \pm 0.13$ & $19.15 \pm 0.21$ & $7.74$ \\
Conv      & $2.88 \pm 0.06$ & $19.18 \pm 0.26$ & $6.85$ \\
BN        & $2.74 \pm 0.05$ & $19.18 \pm 0.34$ & $4.23$ \\
1st Conv  & $2.81 \pm 0.07$ & $19.39 \pm 0.25$ & $\mathbf{3.36}$ \\
1st BN    & $\mathbf{2.72} \pm 0.12$ & $\mathbf{18.80} \pm 0.34$ & $3.37$ \\
\bottomrule
\end{tabular}
\vspace{2pt}
\caption{
Ablation study on different weight sharing strategies.
}
\label{table:ablation}
\end{table}

\begin{figure}
    \centering
    \begin{tabular}{c}
    \includegraphics[width=0.97\linewidth]{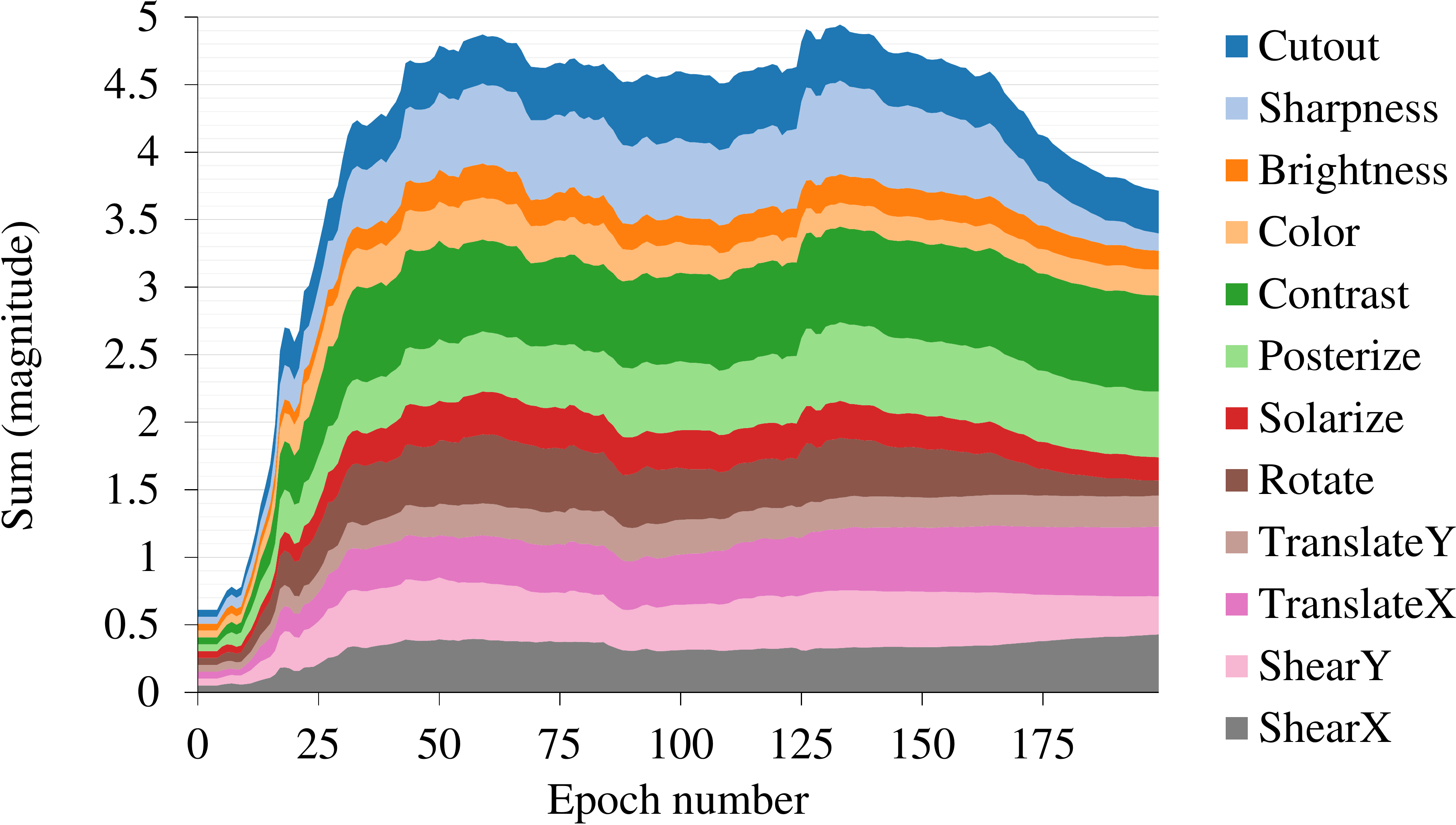} \\
    (a) Operation magnitudes \\
    \includegraphics[width=0.97\linewidth]{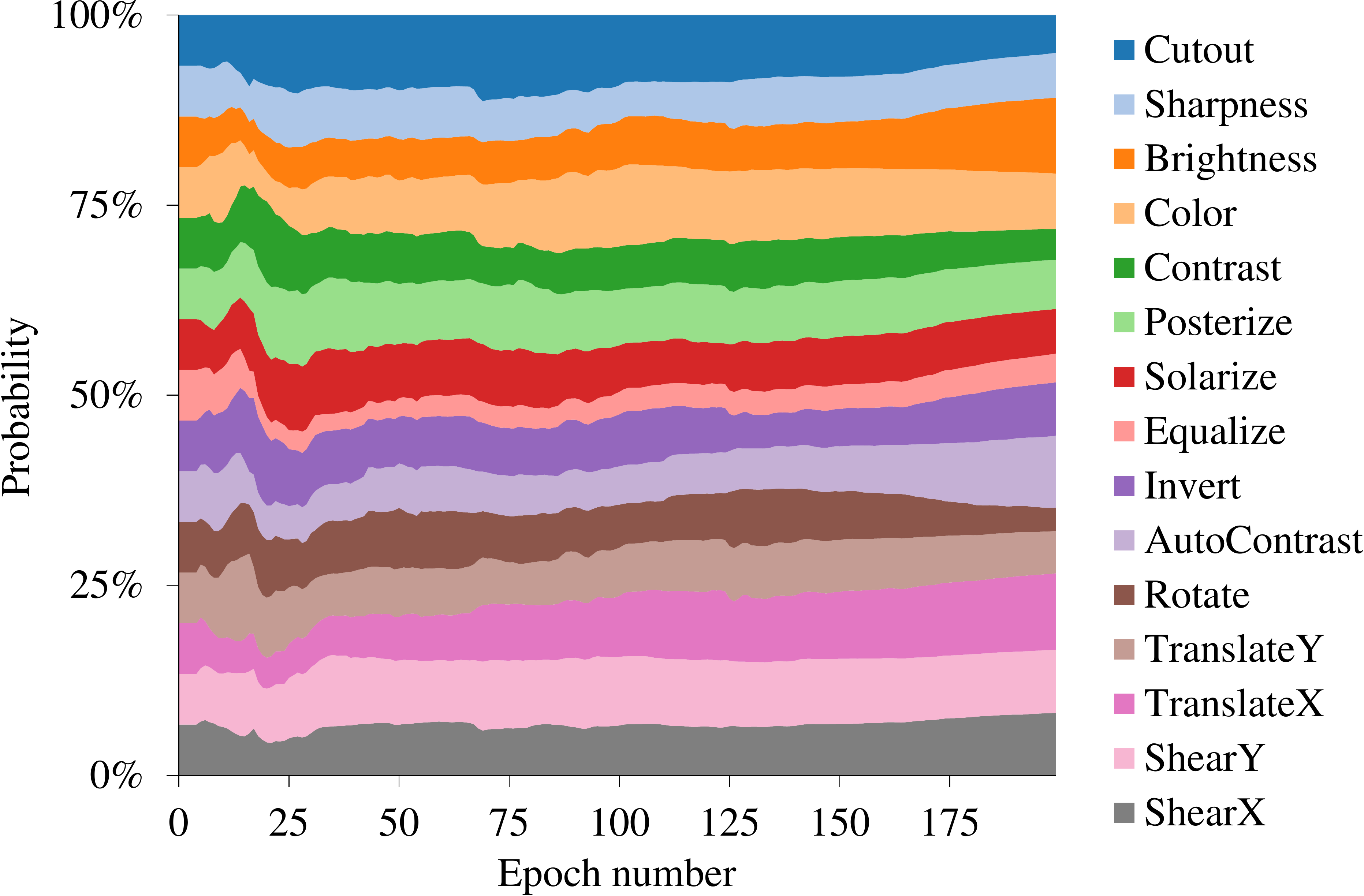} \\
    (b) Normalized probabilities
    \end{tabular}
    \caption{
    The learned hyperparameter schedule of data augmentation. (a) In terms of the operation magnitudes, their sum is rapidly increased initially and slowly decreased at the end. (b) Our schedule is smoothly-varying in terms of the probabilities. The ratio of the largest to the smallest probability is about $5$ during the training.
    }
    \label{fig:schedule}
\end{figure}

\noindent\textbf{Weight Sharing Strategy.}
We first conducted an ablation study on different weight sharing strategies.
In general, if we share a subset of the trainable layers’ parameters from the perspective of population-based training, it means that we do not associate them with hyper-layers.
We experimented with five weight sharing strategies: Conv+BN, Conv, BN, 1st Conv, and 1st BN. These strategies are named by what layers are added with hyper-layers.
For example, the Conv strategy means that we add HyperConv layers to all the convolutional layers and ignore the other layers.
The 1st BN strategy means that we only add a HyperBN layer to the lowest BN layer.

The intuition behind our weight sharing strategies is as follows.
For the low-level layers, we do not share parameters so that they are enforced to reflect the influence of different augmentation policies.
For the mid- and high-level layers, we share parameters to encourage the population to learn a joint feature representation.

In this study, we split the CIFAR training set into three sets (30k/10k/10k) and denoted them as Train30k/Val0/Val.
The CIFAR test set was not used in this study.
For \textit{policy search}, we applied HBA to train Wide-ResNet-28-10 (WRN-28-10)~\cite{zagoruyko2016wide} on the Train30k/Val0 split to find policies.
For \textit{policy evaluation}, we used the searched policies to train networks on the Train30k+Val0 set and evaluated them on the Val set.

Table~\ref{table:ablation} shows the comparison results between these five strategies.
We can see that adding fewer hyper-layers reduces search time and achieves slightly better performance across different datasets, showing that weight sharing in population-based training is helpful in the manner of model cooperation.
We also experimented with WRN-40-2 and obtained similar results, which are included in the supplementary material.
Based on the ablation results, we employ the 1st BN strategy in the subsequent experiments unless otherwise mentioned.

\noindent\textbf{Standard Deviation $\sigmahp$ for exploration.}
Next, we analyzed $\sigmahp$ in Equation~\ref{eq:sgd_w}, representing the range of exploration.
Similar to the previous experiment, we train WRN-28-10 on CIFAR-10.
Three values of $\sigma$ are experimented: $0.5$, $1.0$, and $2.0$, which yield Val set errors of $3.43\pm0.09\%$, $3.30\pm0.10\%$, and $3.38\pm0.05\%$, respectively. Based on the results, we used $\sigma=1.0$ in all of our experiments.

\begin{table}
\centering
\begin{tabular}{l|ccc}
\toprule
Method & CIFAR-10 & SVHN & ImageNet \\
\midrule
AA~\cite{cubuk2019autoaugment} & 5000 & 1000 & 15000 \\
PBA~\cite{ho2019population} & 5 & 1 & - \\
FAA~\cite{lim2019fast} & 3.5 & 1.5 & 450 \\
OHLAA~\cite{lin2019online} & 83.4 & - & 625 \\
AdvAA~\cite{zhang2019adversarial} & - & - & 1280 \\
DADA~\cite{li2020dada} & 0.1 & 0.1 & 1.3 \\
\midrule
\textbf{HBA} & \textbf{0.1} & \textbf{0.1} & \textbf{0.9} \\
\bottomrule
\end{tabular}
\vspace{2pt}
\caption{Comparison of search time in GPU hours. HBA has the least search time and is competitive to DADA.}
\label{table:time}
\end{table}


\begin{table*}
\small
\centering
\begin{tabular}{l|ccccccccc|c}
\toprule
 & Baseline & Cutout & AA & PBA & FAA & OHLAA & RA & AdvAA & DADA & \textbf{HBA} \\
\midrule
\textbf{CIFAR-10} & & & & & & & & & & \\
Wide-ResNet-40-2 & $5.3$ & $4.1$ & $3.7$ & - & $3.6$ & - & - & $3.6$ & - & $3.80 \pm 0.15$ \\ 
Wide-ResNet-28-10 & $3.9$ & $3.1$ & $2.6$ & $2.6$ & $2.7$ & - & $2.7$ & $1.9$ & $2.7$ & $2.63 \pm 0.10$ \\ 
Shake-Shake (26 2x32d) & $3.6$ & $3.0$ & $2.5$ & $2.5$ & $2.7$ & - & - & $2.4$ & $2.7$ & $2.63 \pm 0.16$ \\ 
Shake-Shake (26 2x96d) & $2.9$ & $2.6$ & $2.0$ & $2.0$ & $2.0$ & - & $2.0$ & $1.9$ & $2.0$ & $2.03 \pm 0.10$ \\ 
Shake-Shake (26 2x112d) & $2.8$ & $2.6$ & $1.9$ & $2.0$ & $2.0$ & - & - & $1.8$ & $2.0$ & $1.97 \pm 0.05$ \\ 
PyramidNetN+ShakeDrop & $2.7$ & $2.3$ & $1.5$ & $1.5$ & $1.8$ & - & $1.7$ & $1.4$ & $1.7$ & $1.58 \pm 0.06$ \\ 
\midrule
\textbf{CIFAR-100} & & & & & & & & & & \\
Wide-ResNet-40-2 & $26.0$ & $25.2$ & $20.7$ & - & $20.7$ & - & - & - & $20.9$ & $20.60 \pm 0.37$ \\ 
Wide-ResNet-28-10 & $18.8$ & $18.4$ & $17.1$ & $16.7$ & $17.3$ & - & $16.7$ & $15.5$ & $17.5$ & $17.46 \pm 0.48$ \\ 
Shake-Shake (26 2x96d) & $17.1$ & $16.0$ & $14.3$ & $15.3$ & $14.9$ & - & - & $14.1$ & $15.3$ & $15.35 \pm 0.38$ \\ 
PyramidNetN+ShakeDrop & $14.0$ & $12.2$ & $10.7$ & $10.9$ & $11.9$ & - & - & $10.4$ & $11.2$ & $12.10 \pm 0.14$ \\ 
\midrule
\textbf{SVHN} & & & & & & & & & & \\
Wide-ResNet-28-10      & $1.5$ & $1.3$ & $1.1$ & $1.2$ & $1.1$ & - & $1.0$ & - & $1.2$ & $1.11 \pm 0.02$ \\ 
Shake-Shake (26 2x96d) & $1.4$ & $1.2$ & $1.0$ & $1.1$ & -     & - & -     & - & $1.1$ & $1.05 \pm 0.03$ \\ 
\midrule
\textbf{ImageNet} & & & & & & & & & & \\
ResNet-50 (top-1 error) & $23.7$ & - & $22.4$ & - & $22.4$ & 21.1 & $22.4$ & $20.6$ & $22.5$ & $22.11 \pm 0.10$ \\ 
ResNet-50 (top-5 error) &  $6.9$ & - & $6.2$ & - & $6.3$ & 5.7 &  $6.2$ &  $5.5$ &  $6.5$ &  $6.17 \pm 0.05$ \\ 
\bottomrule
\end{tabular}
\vspace{2pt}
\caption{
The test set error rates $(\%)$ on CIFAR-10, CIFAR-100, SVHN, and ImageNet.
}
\label{table:c10_c100_svhn_in}
\end{table*}

\subsection{Automated Data Augmentation} \label{4.2}

We compare HBA with standard augmentation (Baseline), Cutout~\cite{devries2017improved}, AutoAugment (AA)~\cite{cubuk2019autoaugment}, Population Based Augmentation (PBA)~\cite{ho2019population}, Fast AutoAugment (FAA)~\cite{lim2019fast}, OHL-Auto-Aug (OHLAA)~\cite{lin2019online}, RandAugment (RA)~\cite{cubuk2019randaugment}, Adversarial AutoAugment (AdvAA)~\cite{zhang2019adversarial}, and Differentiable Automatic Data Augmentation (DADA)~\cite{li2020dada}.
Except for Baseline and Cutout, all the methods adopt an AutoAugment-like search space.

\noindent\textbf{Settings.}
We evaluate HBA on four image classification datasets: CIFAR-10~\cite{krizhevsky2009learning}, CIFAR-100~\cite{krizhevsky2009learning}, SVHN~\cite{netzer2011reading}, and ImageNet~\cite{deng2009imagenet}.
For each dataset, we randomly sampled a subset of 10{,}000 images from the original training set as the validation set $D_V$ for HBA.
Following the experimental setting of~\cite{cubuk2019autoaugment, ho2019population, lim2019fast}, we randomly sampled a subset from the remaining images to create a reduced version of the training set.
For \textit{policy search}, we applied HBA to train a WRN-40-2 network on the reduced datasets.
For \textit{policy evaluation}, we evaluated the performance of a model $M$ by using the searched policy to train $M$ on the original training set and measuring the accuracy on the test set.
Please see the supplementary material for the detailed settings.

\noindent\textbf{CIFAR-10 and CIFAR-100.}
The CIFAR-10 and CIFAR-100 datasets consist of 60{,}000 natural images, with a size of 32x32. The training and test sets have 50{,}000 and 10{,}000 images, respectively.
We applied our augmentation policy, baseline, and Cutout (with 16x16 pixels) in sequence to each training image.
The baseline augmentation is defined as the following operations in sequence: standardization, horizontal flipping, and random cropping.
We evaluated our searched policies with WRN-40-2, WRN-28-10, Shake-Shake~\cite{gastaldi2017shake}, and PyramidNet+ShakeDrop~\cite{yamada2019shakedrop}.
As shown in Table~\ref{table:time} and~\ref{table:c10_c100_svhn_in}, HBA significantly improves the performance over the baseline and Cutout, while achieving competitive accuracy with the compared methods.
HBA only takes $0.2$ GPU hours on the Reduced CIFAR-10 for policy search, which is an order of magnitude faster than PBA (population-based) and is comparable to DADA (gradient-based).
In Figure~\ref{fig:schedule}, we visualize the discovered augmentation policy with WRN-40-20 on the CIFAR-10 dataset.

\noindent\textbf{SVHN.}
The SVHN dataset consists of 73{,}257 training images (called core training set), 531{,}131 additional training images, and 26{,}032 test images.
We applied our augmentation policy, baseline, and Cutout (with 20x20 pixels) in sequence to each training image.
The baseline augmentation applied standardization only.
We evaluated our searched policies with WRN-40-2, WRN-28-10, and Shake-Shake.
As shown in Table~\ref{table:time} and~\ref{table:c10_c100_svhn_in}, HBA achieves competitive accuracy and slightly outperforms PBA and DADA.

\noindent\textbf{ImageNet.}
The ImageNet dataset has a training set of about 1.2M images and a validation set of 50{,}000 images.
We applied our augmentation policy and then the baseline augmentation to each training image.
Following~\cite{lim2019fast}, the baseline augmentation applied random resized crop, horizontal flip, color jittering, PCA jittering, and standardization in sequence.
We evaluated our searched policies with ResNet-50~\cite{he2016deep}.
As shown in Table~\ref{table:time} and~\ref{table:c10_c100_svhn_in}, HBA is at least three orders of magnitude faster than all the compared methods except DADA.
Compared with DADA, HBA achieves higher accuracy.

\begin{table*}
\centering
\begin{tabular}{llcllr}
\toprule
\multicolumn{3}{c}{Policy Search}                                                & \multicolumn{3}{c}{Policy Evaluation}            \\
\cmidrule(lr){1-3} \cmidrule(lr){4-6}
Dataset                       & Model                      & Time (GPU hours)                 & Dataset & Model                      & Test Error (\%) \\ 
\midrule
\multirow{4}{*}{Reduced CIFAR-10} & \multirow{4}{*}{WRN-40-2}  & \multirow{4}{*}{$0.1$} & CIFAR-10    & \multirow{2}{*}{WRN-40-2}  & $3.78 \pm 0.15$ \\
           &                            &                      & CIFAR-100   &                            & $20.58 \pm 0.37$ \\
           &                            &                      & CIFAR-10    & \multirow{2}{*}{WRN-28-10} & $2.63 \pm 0.10$  \\
           &                            &                      & CIFAR-100   &                            & $17.46 \pm 0.48$ \\ 
\midrule
CIFAR-10   & \multirow{2}{*}{WRN-40-2}  & \multirow{2}{*}{$1.2 \uparrow$} & CIFAR-10    & \multirow{2}{*}{WRN-40-2}  & $3.63 \pm 0.07 \downarrow$      \\
CIFAR-100  &                            &                      & CIFAR-100   &                            & $20.80 \pm 0.28 \uparrow$       \\
CIFAR-10   & \multirow{2}{*}{WRN-28-10} & \multirow{2}{*}{$4.5 \uparrow$} & CIFAR-10    & \multirow{2}{*}{WRN-28-10} & $2.59 \pm 0.09 \downarrow$      \\
CIFAR-100  &                            &                      & CIFAR-100   &                            & $16.87 \pm 0.22 \downarrow$     \\
\bottomrule
\end{tabular}
\vspace{2pt}
\caption{
Comparison between policy search on the proxy task (top half) and the target tasks (bottom half).
}
\label{table:proxy_vs_target}
\end{table*}

\noindent\textbf{Summary.} Table~\ref{table:time} summarizes the comparison of search time. We can see that HBA has the least search time.
Besides, we also compare with RandAugment (RA), which does not report its search time in its paper.
RA performs a grid search, and its search time (in GPU hours) can be written as $NT$ where $N$ is the number of grid samples, and $T$ is the GPU hours of a single training run.
$N$ is $15$ in its CIFAR-10 experiment.
HBA's search time is about $1.6T$, where $1.6$ is obtained from measuring the ratio of "HBA's training time" to "the typical training time," where both training is conducted on CIFAR-10 with WRN-40-2.
Based on the above analysis, we can see that HBA requires much less GPU hours than RA.

\noindent\textbf{HBA versus DADA.}
In terms of search time, HBA is comparable to DADA, which is the current state-of-the-art.
Comparing with DADA on the accuracy, HBA achieves $3$ wins, $9$ ties, and $1$ loss, as shown in Table~\ref{table:c10_c100_svhn_in}.

\noindent\textbf{Policy Search on the Target Tasks.}
In the previous experiments (Table~\ref{table:time} and~\ref{table:c10_c100_svhn_in}), we applied HBA on a \textit{reduced} task, which trains a smaller network (WRN-40-2) on a reduced dataset.
Here, we compared with the performance of policy searched on the \textit{target} tasks, which trains the target network on the full dataset to find the augmentation policy.
We consider training a model $M \in \{\text{WRN-40-2}, \text{WRN-28-10}\}$ on a dataset $D \in \{\text{CIFAR-10}, \text{CIFAR-100}\}$, giving us four target tasks.
Table~\ref{table:proxy_vs_target} shows that training on the target tasks requires longer search times while achieving slightly better performances (in three out of four cases), which is similar to the findings in RA~\cite{cubuk2019randaugment} and DADA~\cite{li2020dada}.
These results demonstrate that although HBA is designed to pursue efficiency, it can trade off search time for accuracy by searching on the target tasks.

\noindent\textbf{HBA versus RandAugment.}
Although HBA requires fewer GPU hours than RA, HBA does not outperform RA in terms of accuracy.
One may still prefer RA over HBA because of the extreme simplicity of grid search.
We note that the AutoAugment search space is so well-designed that a reduced grid search could suffice.
To address this concern and justify HBA's effectiveness, our next experiment considers a different search space.

\begin{table}
\small
\centering
\begin{tabular}{c|ccc}
\toprule
    & Val. Loss & Test Loss & Test Error $(\%)$ \\
\midrule
GS & $0.794$ & $0.809$ & - \\
RS & $0.921$ & $0.752$ & - \\
STN & $0.579\pm 0.005$ & $0.582\pm 0.004$ & $19.86 \pm 0.21$ \\
\textbf{HBA} & $\mathbf{0.548\pm 0.005}$ & $\mathbf{0.556\pm 0.005}$ & $\mathbf{18.76 \pm 0.11}$ \\
\bottomrule
\end{tabular}
\vspace{1pt}
\caption{Comparison with grid search (GS), random search (RS), and STN on tuning regularization hyperparameters.}
\label{table:stn}
\vspace{-10pt}
\end{table}

\subsection{Comparison with STN}
We followed the experiment settings of STN~\cite{mackay2019self} and applied HBA to simultaneously train an AlexNet~\cite{krizhevsky2012imagenet} on CIFAR-10 and tune regularization hyperparameters.
Here, the search space consists of $15$ regularization hyperparameters: $7$ for data augmentation and $8$ for dropout.
We employed the Conv strategy to construct our hypernetwork for the AlexNet model.
As shown in Table~\ref{table:stn},
both gradient-based methods outperform grid search and random search.
Besides, HBA performs better than STN, showing the effectiveness of our weight sharing strategy.


\section{Conclusion}

We proposed HBA, a gradient-based method for automated data augmentation.
HBA employs a hypernetwork to train a continuous population of models and uses gradient descent to tune augmentation hyperparameters.
Our weight sharing strategy improved both the search speed and accuracy.
Future directions include applying HBA to different domains such as medical images and exploring hybrid algorithms that interpolate between PBT and HBA.

{
\bibliographystyle{ieee_fullname}
\bibliography{egbib}

\begin{thebibliography}{10}\itemsep=-1pt

\bibitem{bender2018understanding}
Gabriel Bender, Pieter-Jan Kindermans, Barret Zoph, Vijay Vasudevan, and Quoc
  Le.
\newblock Understanding and simplifying one-shot architecture search.
\newblock In {\em ICML}, 2018.

\bibitem{berthelot2019mixmatch}
David Berthelot, Nicholas Carlini, Ian Goodfellow, Nicolas Papernot, Avital
  Oliver, and Colin~A Raffel.
\newblock {MixMatch}: A holistic approach to semi-supervised learning.
\newblock In {\em Advances in Neural Information Processing Systems}, pages
  5050--5060, 2019.

\bibitem{brock2017smash}
Andrew Brock, Theodore Lim, James~M Ritchie, and Nick Weston.
\newblock {SMASH}: one-shot model architecture search through hypernetworks.
\newblock In {\em ICLR}, 2018.

\bibitem{chen2020simple}
Ting Chen, Simon Kornblith, Mohammad Norouzi, and Geoffrey Hinton.
\newblock A simple framework for contrastive learning of visual
  representations.
\newblock In {\em ICML}, 2020.

\bibitem{cubuk2019autoaugment}
Ekin~D Cubuk, Barret Zoph, Dandelion Mane, Vijay Vasudevan, and Quoc~V Le.
\newblock {AutoAugment}: Learning augmentation strategies from data.
\newblock In {\em CVPR}, 2019.

\bibitem{cubuk2019randaugment}
Ekin~D Cubuk, Barret Zoph, Jonathon Shlens, and Quoc~V Le.
\newblock {RandAugment}: Practical automated data augmentation with a reduced
  search space.
\newblock In {\em NeurIPS}, 2020.

\bibitem{deng2009imagenet}
Jia Deng, Wei Dong, Richard Socher, Li-Jia Li, Kai Li, and Li Fei-Fei.
\newblock {ImageNet}: A large-scale hierarchical image database.
\newblock In {\em CVPR}, 2009.

\bibitem{devries2017improved}
Terrance DeVries and Graham~W Taylor.
\newblock Improved regularization of convolutional neural networks with cutout.
\newblock {\em arXiv preprint arXiv:1708.04552}, 2017.

\bibitem{gastaldi2017shake}
Xavier Gastaldi.
\newblock Shake-shake regularization.
\newblock In {\em ICLR}, 2017.

\bibitem{ha2017hypernetworks}
David Ha, Andrew Dai, and Quoc~V Le.
\newblock Hypernetworks.
\newblock In {\em ICLR}, 2017.

\bibitem{he2016deep}
Kaiming He, Xiangyu Zhang, Shaoqing Ren, and Jian Sun.
\newblock Deep residual learning for image recognition.
\newblock In {\em CVPR}, 2016.

\bibitem{ho2019population}
Daniel Ho, Eric Liang, Ion Stoica, Pieter Abbeel, and Xi Chen.
\newblock Population based augmentation: Efficient learning of augmentation
  policy schedules.
\newblock In {\em ICML}, 2019.

\bibitem{ioffe2015batch}
Sergey Ioffe and Christian Szegedy.
\newblock Batch normalization: Accelerating deep network training by reducing
  internal covariate shift.
\newblock In {\em ICML}, 2015.

\bibitem{jaderberg2017population}
Max Jaderberg, Valentin Dalibard, Simon Osindero, Wojciech~M Czarnecki, Jeff
  Donahue, Ali Razavi, Oriol Vinyals, Tim Green, Iain Dunning, Karen Simonyan,
  et~al.
\newblock Population based training of neural networks.
\newblock {\em arXiv preprint arXiv:1711.09846}, 2017.

\bibitem{kingma2014adam}
Diederick~P Kingma and Jimmy Ba.
\newblock Adam: A method for stochastic optimization.
\newblock In {\em ICLR}, 2015.

\bibitem{kostrikov2020image}
Ilya Kostrikov, Denis Yarats, and Rob Fergus.
\newblock Image augmentation is all you need: Regularizing deep reinforcement
  learning from pixels.
\newblock In {\em ICLR}, 2021.

\bibitem{krizhevsky2009learning}
Alex Krizhevsky and Geoffrey Hinton.
\newblock Learning multiple layers of features from tiny images.
\newblock {\em Technical Report}, 2009.

\bibitem{krizhevsky2012imagenet}
Alex Krizhevsky, Ilya Sutskever, and Geoffrey~E Hinton.
\newblock Imagenet classification with deep convolutional neural networks.
\newblock In {\em Advances in neural information processing systems}, 2012.

\bibitem{li2020dada}
Yonggang Li, Guosheng Hu, Yongtao Wang, Timothy~M. Hospedales, Neil~Martin
  Robertson, and Yongxin Yang.
\newblock {DADA:} differentiable automatic data augmentation.
\newblock In {\em ECCV}, 2020.

\bibitem{lim2019fast}
Sungbin Lim, Ildoo Kim, Taesup Kim, Chiheon Kim, and Sungwoong Kim.
\newblock Fast {AutoAugment}.
\newblock In {\em NeurIPS}, 2019.

\bibitem{lin2019online}
Chen Lin, Minghao Guo, Chuming Li, Xin Yuan, Wei Wu, Junjie Yan, Dahua Lin, and
  Wanli Ouyang.
\newblock Online hyper-parameter learning for auto-augmentation strategy.
\newblock In {\em ICCV}, 2019.

\bibitem{liu2019darts}
Hanxiao Liu, Karen Simonyan, and Yiming Yang.
\newblock {DARTS}: Differentiable architecture search.
\newblock In {\em ICLR}, 2019.

\bibitem{mackay2019self}
Matthew MacKay, Paul Vicol, Jon Lorraine, David Duvenaud, and Roger Grosse.
\newblock Self-tuning networks: Bilevel optimization of hyperparameters using
  structured best-response functions.
\newblock In {\em ICLR}, 2019.

\bibitem{netzer2011reading}
Yuval Netzer, Tao Wang, Adam Coates, Alessandro Bissacco, Bo Wu, and Andrew~Y.
  Ng.
\newblock Reading digits in natural images with unsupervised feature learning.
\newblock In {\em NIPS Workshop on Deep Learning and Unsupervised Feature
  Learning 2011}, 2011.

\bibitem{paszke2019pytorch}
Adam Paszke, Sam Gross, Francisco Massa, Adam Lerer, James Bradbury, Gregory
  Chanan, Trevor Killeen, Zeming Lin, Natalia Gimelshein, Luca Antiga, et~al.
\newblock {PyTorch}: An imperative style, high-performance deep learning
  library.
\newblock In {\em Advances in Neural Information Processing Systems}, 2019.

\bibitem{pham2018efficient}
Hieu Pham, Melody~Y Guan, Barret Zoph, Quoc~V Le, and Jeff Dean.
\newblock Efficient neural architecture search via parameter sharing.
\newblock In {\em ICML}, 2018.

\bibitem{shorten2019survey}
Connor Shorten and Taghi~M Khoshgoftaar.
\newblock A survey on image data augmentation for deep learning.
\newblock {\em Journal of Big Data}, 2019.

\bibitem{srivastava2014dropout}
Nitish Srivastava, Geoffrey Hinton, Alex Krizhevsky, Ilya Sutskever, and Ruslan
  Salakhutdinov.
\newblock Dropout: a simple way to prevent neural networks from overfitting.
\newblock {\em JMLR}, 2014.

\bibitem{xie2019unsupervised}
Qizhe Xie, Zihang Dai, Eduard Hovy, Minh-Thang Luong, and Quoc~V Le.
\newblock Unsupervised data augmentation for consistency training.
\newblock In {\em NeurIPS}, 2020.

\bibitem{yamada2019shakedrop}
Yoshihiro Yamada, Masakazu Iwamura, Takuya Akiba, and Koichi Kise.
\newblock Shakedrop regularization for deep residual learning.
\newblock {\em IEEE Access}, 2019.

\bibitem{yu2020hyper}
Tong Yu and Hong Zhu.
\newblock Hyper-parameter optimization: A review of algorithms and
  applications.
\newblock {\em arXiv preprint arXiv:2003.05689}, 2020.

\bibitem{yun2019cutmix}
Sangdoo Yun, Dongyoon Han, Seong~Joon Oh, Sanghyuk Chun, Junsuk Choe, and
  Youngjoon Yoo.
\newblock {CutMix}: Regularization strategy to train strong classifiers with
  localizable features.
\newblock In {\em ICCV}, 2019.

\bibitem{zagoruyko2016wide}
Sergey Zagoruyko and Nikos Komodakis.
\newblock Wide residual networks.
\newblock In {\em BMVC}, 2016.

\bibitem{zhang2017mixup}
Hongyi Zhang, Moustapha Cisse, Yann~N Dauphin, and David Lopez-Paz.
\newblock mixup: Beyond empirical risk minimization.
\newblock In {\em ICLR}, 2018.

\bibitem{zhang2019adversarial}
Xinyu Zhang, Qiang Wang, Jian Zhang, and Zhao Zhong.
\newblock Adversarial {AutoAugment}.
\newblock In {\em ICLR}, 2020.

\bibitem{zoph2017neural}
Barret Zoph and Quoc~V Le.
\newblock Neural architecture search with reinforcement learning.
\newblock In {\em ICLR}, 2017.

\end{thebibliography}
}

\begin{appendices}
\section{Search space}

\noindent\textbf{Elementary Image Operations.}
Table~\ref{table:operation} shows the list of augmentation operations used in our experiments except for Section 4.3.
When applying an operation $op \in \{\text{ShearX}, \text{ShearY}, \text{TranslateX}, \text{TranslateY}, \text{Rotate}\}$ to an input image, we randomly negate the sampled magnitude with a probability of $0.5$.

\noindent\textbf{Augmentation Function.}
Following PBA~\cite{ho2019population}, we use a categorical distribution $K\sim~$\{0: $p$ = 0.2, 1: $p$ = 0.3, 2: $p$ = 0.5\} as the number of operations to bee applied.
Please refer to the Algorithm 1 in the PBA paper for the augmentation function used for both PBA and our HBA.

\noindent\textbf{Initialization.}
For each augmentation hyperparameter, we initialize its value by $0.95M_\text{min}+0.05M_\text{max}$, where $[M_\text{min}, M_\text{max}]$ is the magnitude range of the hyperparameter.

\noindent\textbf{Implementation Details.}
Following PBA~\cite{ho2019population}, our search space consists of $15$ operations, where each has two copies.
We note that, unlike PBA that discretizes continuous hyperparameters, we treat all hyperparameters (both probabilities and magnitudes) as continuous variables.
Following STN~\cite{mackay2019self}, we define the hyperparameters $\hp$ as the unbounded version of the operation magnitudes and probabilities by mapping a range of magnitude $[M_\text{min}, M_\text{max}]$ to $[-\infty, \infty]$ through a logit function (the inverse of the sigmoid function).
The magnitude range of each operation is shown in Table~\ref{table:operation}, and the probability range of each operation is $[0, 1]$.

\begin{table*}[h]
\caption{List of augmentation operations.}
\label{table:operation}
\centering
\begin{tabular}{ccc}
\toprule
Operation & Range of Magnitude & Unit of Magnitude       \\
\midrule
ShearX         & {[}0, 0.3{]}       & -          \\
ShearY         & {[}0, 0.3{]}       & -          \\
TranslateX     & {[}0, 0.45{]}      & Image size \\
TranslateY     & {[}0, 0.45{]}      & Image size \\
Rotate         & {[}0, 30{]}        & Degree     \\
AutoContrast   & None               & -          \\
Invert         & None               & -          \\
Equalize       & None               & -          \\
Solarize       & {[}0, 255{]}       & -          \\
Posterize      & {[}0, 8{]}         & Bit        \\
Contrast       & {[}0.1, 1.9{]}     & -          \\
Color          & {[}0.1, 1.9{]}     & -          \\
Brightness     & {[}0.1, 1.9{]}     & -          \\
Sharpness      & {[}0.1, 1.9{]}     & -          \\
Cutout         & {[}0, 0.2{]}       & Image size \\
\bottomrule
\end{tabular}
\end{table*}
\section{Population-Based Training Procedure}

Our population-based training procedure can be expressed in the format of PBT~\cite{jaderberg2017population} as follows:

\noindent\textbf{Step}: each model runs an SGD step with a batch size of $1$.

\noindent\textbf{Eval}: we evaluate each model on a validation set by cross entropy loss.

\noindent\textbf{Ready}: each model goes through the exploit-and-explore process every $\trainstepN$ steps.

\noindent\textbf{Exploit}: each model clones the weights and hyperparameters of the best performing model.

\noindent\textbf{Explore}: for each model, we perturb the value of its parameters and hyperparameters.

\noindent
\begin{minipage}{0.47\textwidth}
\begin{algorithm}[H]
\algsetup{linenosize=\small}
\caption{Our population-based training procedure.}
\begin{algorithmic}[1]
    \STATE {\bfseries Input:} population size $\mN$, number of steps $\outerN$ and $\trainstepN$, training set $\trainset$, augmentation policy $\aug$, neural network $\model$, learning rate $\trainLR$, validation set $\valset$, Gaussian sigma $\sigmahp$ and $\sigmaw$.
    \STATE Initialize $\{ \w_\mi \}_{\mi=1}^{\mN}$ and $\{ \hp_\mi \}_{\mi=1}^{\mN}$
    \FOR{$j=1$ {\bfseries to} $\outerN$}
        \FOR{$\mi=1$ {\bfseries to} $\mN$ (synchronously in parallel)}
            \FOR{$t=1$ {\bfseries to} $\trainstepN$}
                \STATE // \textit{Step}
                \STATE Sample $(\inputimg,\gtlabel)$ from $\trainset$
                \STATE $\w_\mi = \w_\mi - \trainLR \nabla_{\w} \loss(\model(\aug(\inputimg;\hp_\mi);\w_\mi), \gtlabel)$
            \ENDFOR
        \ENDFOR
        \STATE // \textit{Eval}
        \STATE $\mk = \arg\min_{\mi\in\{1,...,\mN\}}$ \\
        $~~~~~~~\sum_{(\inputimgval,\gtlabelval)\in \valset}\loss(\model(\inputimgval;\w_\mi), \gtlabelval)$
        \STATE // \textit{Exploit}
        \FOR{$\mi=1$ {\bfseries to} $\mN$}
            \STATE $\hp_\mi=\hp_\mk$
            \STATE $\w_\mi=\w_\mk$
        \ENDFOR
        \STATE // \textit{Explore}
        \FOR{$\mi=1$ {\bfseries to} $\mN$}
            \STATE $\hp_\mi=\hp_\mi+\noisehp_\mi$ where $\noisehp_\mi \sim \gaussian(0,\sigmahp)$ 
            \STATE $\w_\mi=\w_\mi+\noisew_\mi$ where $\noisew_\mi \sim \gaussian(0,\sigmaw)$      
        \ENDFOR
    \ENDFOR
    \STATE {\bfseries Output:} $\w_\mk$
\end{algorithmic}
\end{algorithm}
\end{minipage}
\section{Implementation Details of Hyper-layers}

\subsection{Hyper-Linear Layer}
Let $\linear$ be a linear layer.
We have
\begin{align}
    \outputvec = \linear(\inputvec; \weightmatrix) = \weightmatrix\inputvec,
\end{align}
where $\inputvec \in \Real^{\indim}$, $\outputvec \in \Real^{\outdim}$, and the weight matrix $\weightmatrix \in \Real^{\outdim \times \indim}$ is the parameters of the linear layer.
Given a linear layer $\linear$, we define its corresponding Hyper-linear layer $\hyperlinear$ as a linear function that maps hyperparameters $\hp \in \Real^{\hpN}$ to the weight matrix $\weightmatrix \in \Real^{\outdim \times \indim}$ of the linear layer $\linear$.
We can decompose the hyperlinear layer $\hyperlinear$ into a set linear functions $\{ \subhyperlinear_i(\hp) \}_{i=1}^{\outdim}$ that each function $\subhyperlinear_i(\hp)$ outputs the transpose of the i-th row of $\weightmatrix$. $\hyperlinear$ can be expressed as
\begin{align}
    \weightmatrix = \hyperlinear(\hp) &= [\subhyperlinear_1(\hp), ..., \subhyperlinear_{\outdim}(\hp)]^T \\
    \text{and~~} \subhyperlinear_{i}(\hp) &= \subhyperlinearbias_i + \subhyperlinearweight_i \hp,
\end{align}
where $\subhyperlinearweight_i \in \Real^{\indim \times \hpN}$ and $\subhyperlinearbias_i \in \Real^{\indim}$.
A Hyper-linear layer parameterized by $\{\subhyperlinearweight_i\}_{i=1}^{\outdim}$ and $\{\subhyperlinearbias_i\}_{i=1}^{\outdim}$ requires $\hpN\indim\outdim$ and $\indim\outdim$ parameters, respectively.
Due to its prohibitively huge memory consumption, following Self-Tuning Network (STN)~\cite{mackay2019self}, we assume $\subhyperlinearweight_i$ is a rank-$1$ matrix to greatly reduce the number of parameters of the Hyper-linear layer.
Specifically, we define $\subhyperlinearweight_i = \uvec_i \vvec_i^T$ where $\uvec_i \in \Real^{\indim}$ and $\vvec_i \in \Real^{\hpN}$.
By doing so, the number of parameters of the Hyper-linear layer is reduced to $(\hpN+\cin)\cout + \cin\cout$.
$\subhyperlinear_{i}(\hp)$ is written as
\begin{align}
    \subhyperlinear_{i}(\hp) & = \subhyperlinearbias_i + \subhyperlinearweight_i \hp = \subhyperlinearbias_i + \uvec_i \vvec_i^T \hp = \subhyperlinearbias_i + (\vvec_i^T \hp) \uvec_i.
\end{align}

The Hyper-linear layer $\hyperlinear(\hp)$ can then be expressed as
\begin{align}
\weightmatrix = \hyperlinear(\hp) &= [\subhyperlinear_1(\hp), ..., \subhyperlinear_{\cout}(\hp)]^T \\
&= \hyperlinearbias + \diag{\hyperlinearV\hp}\hyperlinearU
\end{align}
where
$\hyperlinearbias = [\subhyperlinearbias_1, ..., \subhyperlinearbias_{\cout}]^T \in \Real^{\cout \times \cin}$, 
$\hyperlinearV = [\vvec_1, ..., \vvec_{\cout}]^T \in \Real^{\cout \times \hpN}$,
$\hyperlinearU = [\uvec_1, ..., \uvec_{\cout}]^T \in \Real^{\cout \times \cin}$, 
and $\diag{\cdot}$ turns a vector into a diagonal matrix.
In particular, $\weightmatrix$, $\hyperlinearbias$, and $\hyperlinearU$ have the same matrix size.
Let $\hyperlinearbias$, $\hyperlinearU$, and $\hyperlinearV$ be denoted by $\hnetw_0$, $\hnetw_U$, and $\hnetw_V$, respectively.
The hyper-linear layer can be expressed in a general form as
\begin{align}
    \weightmatrix = \hyperlinear(\hp;\hnetw_0,\hnetw_U,\hnetw_V) = \hnetw_0 + \diag{\hnetw_V \hp}\hnetw_{U}.
    \label{eq:linear_w}
\end{align}
Consider a linear layer with a bias $\bm{b}$, the bias part of the hyperLinear layer can be additionally defined in a similar way as
\begin{align}
    \bias = \hyperbias(\hp; \hyperbiasbias, \hyperbiasU, \hyperbiasV) = \hyperbiasbias + \diag{\hyperbiasV\hp}\hyperbiasU
    \label{eq:linear_b}
\end{align}
where $\hyperbiasbias \in \Real^{\cout}$, $\hyperbiasV \in \Real^{\cout \times \hpN}$, and $\hyperbiasU \in \Real^{\cout}$.


\subsection{Hyper-Conv Layer}
A linear layer can be interpreted as a $1\times 1$ convolutional layer by (1) viewing the input vector $\inputvec$ as an image with $\cin$ channels and a spatial size of $1\times 1$, and (2) viewing the weight matrix $\weightmatrix$ as the set of $1\times 1$ convolutional filters, where each row of $\weightmatrix$ is a filter.
Therefore, we can define the hyper-layer of a $1\times 1$ convolutional layer in the same way as the hyper-linear layer.
We follow the definition of the hyper-linear layer (Equation~\ref{eq:linear_w}), and define the hyper-conv1x1 layer as
\begin{align}
    \convONEweight &= \hyperconvONE(\hp;\hyperconvONEbias,\hyperconvONEU, \hyperconvONEV) \\
    &= \hyperconvONEbias + \diag{\hyperconvONEV\hp}\hyperconvONEU, 
\end{align} 
where $\convONEweight, \hyperconvONEbias, \hyperconvONEU \in \Real^{\cout\times\cin}$ are three sets of $1\times 1$ filters.
$\cin$ and $\cout$ are the number of the input and output channels, respectively.
$\diag{\hyperconvONEV\hp}\hyperconvONEU$ can be interpreted as a filter-wise scaling of $\hyperconvONEU$, in which the j-th filter weights of $\hyperconvONEU$ are scaled by the j-th element of $\hyperconvONEV\hp$.
In general, the hyper-conv layer for a $k\times k$ convolutional layer can be defined with the  hyper-conv1x1 layer by changing $\hyperconvONEbias$ and $\hyperconvONEU$ from $1\times 1$ to $k\times k$ filters as follows:
\begin{align}
    \convweight = \hyperconv(\hp;\hyperconvbias,\hyperconvU, \hyperconvV) = \hyperconvbias + \diag{\hyperconvV\hp}\hyperconvU, 
\end{align} 
where each row of $\hyperconvbias$ and $\hyperconvU$ corresponds to the parameters of a $k\times k$ filter.
Specifically, $\convweight, \hyperconvbias, \hyperconvU \in \Real^{\cout\times\cin k^2}$, $\hyperconvV \in \Real^{\cout \times \hpN}$.
For a Conv layer with a bias, the bias part of the hyper-conv layer is the same as the hyper-linear one (Equation~\ref{eq:linear_b}).

\subsection{Hyper-BN Layer}
A batch normalization layer has a trainable affine transformation.
Following the design spirit of the hyper-linear and the hyper-conv layer, we denote the affine parameters by $\w_\text{BN}$ and define the hyper-BN layer as 
\begin{align}
    \BNaffine &= \text{hyper-BN}(\hp;\hyperBNbias,\hyperBNU,\hyperBNV) \\
    &= \hyperBNbias + \diag{\hyperBNV \hp}\hyperBNU,
\end{align}
where $\hyperBNbias \in \Real^{2\cout}$, $\hyperBNV \in \Real^{2\cout \times \hpN}$, $\hyperBNU \in \Real^{2\cout}$, and $\cout$ is the number of the output channels. There is a factor $2$ because the affine transformation has $\cout$ scaling parameters and $\cout$ offset parameters.

\section{Hyperparameters}

Table~\ref{table:search_hp} and~\ref{table:eval_hp} show the hyperparameters used in policy search and policy evaluation, respectively.
For the ImageNet dataset, we used a step-decay learning rate schedule that drops by $0.1$ at epoch $90$, $180$, and $240$.

\begin{table*}
\centering
\begin{tabular}{ccccc}
\toprule
Dataset         & \begin{tabular}[c]{@{}c@{}}Reduced\\ CIFAR-10\end{tabular} & \begin{tabular}[c]{@{}c@{}}Reduced\\ SVHN\end{tabular} & \begin{tabular}[c]{@{}c@{}}Reduced\\ ImageNet\end{tabular} & CIFAR-10 / CIFAR-100 \\
\midrule
No. classes & 10 & 10 & 120 & 10 \\
No. training images & 4,000 & 4,000 & 6,000 & 40,000 \\
No. validation images   & 10,000 & 10,000 & 10,000 & 10,000   \\
Model   & WRN-40-2   & WRN-40-2   & WRN-40-2   & WRN-40-2 / WRN-28-10 \\
Input size & 32x32 & 32x32 & 32x32 & 32x32 \\
No. training epoch  & 200& 160& 270& 200  \\
Learning rate $\alpha$ & $0.05$   & $0.01$   & $0.1$   & $0.1$ \\
Learning rate schedule ($\alpha$) & cosine & cosine & step   & cosine   \\
Weight decay& $0.005$   & $0.01$   & $0.001$   & $0.0005$ \\
Batch size  & 128 & 128 & 128   & 128  \\
Learning rate $\alpha'$ & $0.03$   & $0.02$   & $0.007$  & $0.003$ \\
Learning rate schedule ($\alpha'$) & constant & constant & constant & constant   \\
\midrule
Results in the main paper & Table 4 & Table 4 & Table 4 & Table 5 \\
\bottomrule
\end{tabular}
\vspace{4pt}
\caption{Hyperparameters for policy search.}
\label{table:search_hp}
\end{table*}

\begin{table*}
\centering
\begin{tabular}{lllclcc}
\toprule
Dataset   & Model & LR & Schedule & WD & BS & Epoch \\
\midrule
CIFAR-10  & WRN-40-2                & 0.1  & cosine & 0.0005  & 128 & 200   \\
CIFAR-10  & WRN-28-10               & 0.1  & cosine & 0.0005  & 128 & 200   \\
CIFAR-10  & Shake-Shake (26 2x32d)  & 0.01 & cosine & 0.001   & 128 & 1800     \\
CIFAR-10  & Shake-Shake (26 2x96d)  & 0.01 & cosine & 0.001   & 128 & 1800     \\
CIFAR-10  & Shake-Shake (26 2x112d) & 0.01 & cosine & 0.001   & 128 & 1800     \\
CIFAR-10  & PyramidNet+ShakeDrop    & 0.05 & cosine & 0.00005 & 64  & 1800     \\
\midrule
CIFAR-100 & WRN-28-10               & 0.1  & cosine & 0.0005  & 128 & 200   \\
CIFAR-100 & Shake-Shake (26 2x96d)  & 0.01 & cosine & 0.0025  & 128 & 1800     \\
CIFAR-100 & PyramidNet+ShakeDrop    & 0.025 & cosine & 0.0005  & 64  & 1800     \\
\midrule
SVHN      & WRN-28-10               & 0.005 & cosine & 0.001   & 128 & 160     \\
SVHN      & Shake-Shake (26 2x96d)  & 0.01  & cosine & 0.00015 & 128 & 160     \\
\midrule
ImageNet  & ResNet-50               & 0.1   & step & 0.0001       & 256   & 270     \\
\bottomrule
\end{tabular}
\vspace{4pt}
\caption{Hyperparameters for policy evaluation. LR: learning rate. Schedule: learning rate schedule. WD: weight decay. BS: batch size. Epoch: number of training epoch. All the hyperparameters follow the settings in PBA~\cite{ho2019population} except the ones for the ImageNet dataset.
We used $8$ NVIDIA Tesla V100 GPUs to train ResNet-50 on the ImageNet dataset.}
\label{table:eval_hp}
\end{table*}

\section{Ablation study of weight sharing strategies.}
In Table~\ref{table:ablation}, we show the full results of the ablation study on weight sharing strategies. In particular, we experimented with both Wide-ResNet-40-2 and Wide-ResNet-28-10. Their results are consistent: adding fewer hyper-layers reduces search time and achieves slightly better performance.

\begin{table*}
\centering
\begin{tabular}{c|ccc|ccc}
\toprule
             & \multicolumn{3}{c|}{Wide-ResNet-40-2} & \multicolumn{3}{c}{WideResNet-28-10} \\
Hyper-layers & CIFAR-10 & CIFAR-100 & Time  & CIFAR-10 & CIFAR-100 & Time \\
             & Val. Error (\%) & Val. Error (\%) & (GPU Hours) & Val. Error (\%) & Val. Error (\%) & (GPU Hours) \\
\midrule
Conv + BN & $4.01 \pm 0.09$ & $22.66 \pm 0.19$ & $2.03$         & $2.85 \pm 0.13$ & $19.15 \pm 0.21$ & $7.74$ \\
Conv      & $3.94 \pm 0.08$ & $22.71 \pm 0.34$ & $1.53$         & $2.88 \pm 0.06$ & $19.18 \pm 0.26$ & $6.85$ \\
BN        & $3.86 \pm 0.08$ & $22.86 \pm 0.22$ & $1.33$         & $2.74 \pm 0.05$ & $19.18 \pm 0.34$ & $4.23$ \\
1st Conv & $\mathbf{3.85} \pm 0.06$ & $22.66 \pm 0.40$ & $\mathbf{0.86}$         & $2.81 \pm 0.07$ & $19.39 \pm 0.25$ & $\mathbf{3.36}$ \\
1st BN   & $4.02 \pm 0.11$ & $\mathbf{22.20} \pm 0.10$ & $0.87$         & $\mathbf{2.72} \pm 0.12$ & $\mathbf{18.80} \pm 0.34$ & $3.37$ \\
\bottomrule
\end{tabular}
\vspace{2pt}
\caption{
Ablation study on different weight sharing strategies. 
}
\label{table:ablation}
\end{table*}

\end{appendices}

\end{document}